\documentclass{article}

    \usepackage[final]{neurips_2024}

\usepackage[utf8]{inputenc} 
\usepackage[T1]{fontenc}    
\usepackage{hyperref}       
\usepackage{url}            
\usepackage{booktabs}       
\usepackage{amsfonts}       
\usepackage{nicefrac}       
\usepackage{microtype}      
\usepackage{xcolor}         

\usepackage{graphicx}
\usepackage{subfigure}

\usepackage{amsmath}
\usepackage{amssymb}
\usepackage{mathtools}
\usepackage{amsthm}
\usepackage{pifont}
\usepackage{caption}
\usepackage{multirow}
\usepackage{enumitem}

\title{PQA: Zero-shot Protein Question Answering for Free-form Scientific Enquiry with Language Models}

\author{%
  Eli M Carrami\thanks{Corresponding Author: \texttt{eli.carrami@gmail.com}} \\
  London, UK\\
  \And
  Sahand Sharifzadeh \\
  LMU Munich \\
  Munich, Germany \\
}

\begin{document}

\maketitle

\begin{abstract}
Understanding protein structure and function is crucial in biology. However, current computational methods are often task-specific and resource-intensive. To address this, we propose zero-shot Protein Question Answering (PQA), a task designed to answer a wide range of protein-related queries without task-specific training.
The success of PQA hinges on high-quality datasets and robust evaluation strategies, both of which are lacking in current research. Existing datasets suffer from biases, noise, and lack of evolutionary context, while current evaluation methods fail to accurately assess model performance.
We introduce the Pika framework to overcome these limitations. Pika comprises a curated, debiased dataset tailored for PQA and a biochemically relevant benchmarking strategy. We also propose multimodal large language models as a strong baseline for PQA, leveraging their natural language processing and knowledge.
This approach promises a more flexible and efficient way to explore protein properties, advancing protein research. Our comprehensive PQA framework, Pika, including dataset, code, and model checkpoints, is openly accessible on \href{github.com/EMCarrami/Pika}{Github}, promoting wider research in the field.

\end{abstract}

\section{Introduction}

\label{submission}
Proteins, essential to biological functions, are complex macromolecules that perform a myriad of cellular roles determined by their complex structures and interactions. As a polymer of amino acids, drawn from a pool of 20 natural ones, the sequence in which these building blocks are arranged dictates the protein's three-dimensional structure, which is critical for its function. Given the critical role of proteins in both fundamental biology and applied biomedical research, a deep understanding of their structures and functions is crucial. Despite significant advances in deciphering proteins' 3D structures, there remains a pressing need for innovative methodologies to facilitate the computational study of their biochemical and functional properties.

Currently, training individual models tailored to specific tasks are the primary approach to computationally studying the biochemical and functional properties of proteins, requiring extensive data collection and training for each unique task.
For instance, submissions to the Critical Assessment of Protein Function Annotation algorithms (CAFA) aim to predict the functional annotations such as GO terms for new protein sequences, as required by the multi-year challenge \citep{cafa}. Similarly, several other task-specific models have been developed to predict specific biochemical properties of proteins, such as ligand binding \citep{Wei2022DeepLPI} or thermal stability \citep{Blaabjerg2023}.  To address the limitations of current approaches, we propose to unify protein sequence-related enquiries under a more generic task of zero-shot Protein Question Answering (PQA), where there can be free-form inquiries about known or novel protein sequences. 

The development and assessment of effective zero-shot PQA models critically depend on the availability of high-quality \textbf{datasets} and robust \textbf{evaluation strategies}, both of which present significant challenges in the current research landscape; unfortunately, existing datasets available for related tasks often exhibit significant limitations. They frequently demonstrate biases towards specific protein families, overlook the critical evolutionary relationships between proteins, and may contain noisy or unreliable annotations. These shortcomings hinder the development of effective PQA models that can generalize across diverse proteins and accurately answer complex queries.

Similarly, the robust evaluation of PQA models necessitates biochemically-relevant benchmarking strategies that focus on the scientific accuracy of model predictions within the context of the specific questions posed. Previous research efforts have not adequately addressed these challenges. They often neglect the impact of evolutionary relationships on information leakage, which can lead to overestimation of model performance. Moreover, they predominantly rely on metrics like BLEU \citep{papineni2002bleu} and ROUGE \citep{rouge}, which have been consistently shown to be inadequate for assessing the accuracy of scientific statements \citep{Mathur2020}. This lack of suitable evaluation tools further impedes progress in the field.

To address these limitations, we have developed the Pika framework, comprised of a curated and debiased dataset specifically designed for the PQA domain, equipped with scientific question and answer (QA) pairs for instructional training as well as a robust and biochemically relevant benchmarking strategy to enable effective evaluation of PQA models. Alongside, we also propose multimodal large language models (LLMs) as a potential solution to PQA and create evolution-aware splits for assessment of their performance. Besides their capability in processing natural questions, LLMs encapsulate a large body of knowledge which could provide further context for the model, enabling it to perform a more flexible and efficient exploration of protein functionalities, bypassing the need for extensive model training and data collection for individual queries. 

Finally, comparing the performance of multimodal LLMs with various relevant baselines on our evolution-aware data splits, we show that these models are a promising direction for PQA. In particular, we train and evaluate two multimodal protein-text architectures combining the ESM2 \citep{lin2022language} protein language model (PLM) with the Phi-2 LLM \citep{huggingface2023phi2}, showcasing the seamless integration of protein sequence analysis with natural language processing while highlight the shortcomings of these models in particular when dealing with evolutionary distant proteins. Our results suggest that the strategic adoption of this methodology, especially with more advanced LLMs, has the potential to challenge the current state-of-the-art in task-specific models. It is essential to highlight that, our research, leveraging the robust yet modestly scaled Phi-2 LLM, serves as a robust proof-of-concept to demonstrate the potential of this approach. We hope our work fuels further research and opens new avenues towards performant PQA models.

\section{Related Work}
\label{sec:LitPQA}

Since PQA is distinct from other question-answering tasks, existing scientific QA datasets such as ScienceQA \citep{lu2022learn}, PubMedQA \citep{jin2019pubmedqa} or SQuAD \citep{rajpurkar2016squad, rajpurkar2018know} are not suitable for PQA training or evaluation. This is because unlike standard QA tasks where the answer to the question can be inferred using logic or existing knowledge, in PQA the answer to the question must be extracted from cross-modality embeddings via the LLM, therefore necessitating specialized protein-text datasets that annotate protein sequences with relevant labels.

Previous studies on integrating text and protein sequences have primarily leveraged large-scale unstructured biomedical text or knowledge graphs to refine protein representations, facilitating downstream tasks like functional classification and sequence generation (e.g. ProtST~\citep{xu2023protst}, OntoProtein~\citep{zhang2022ontoprotein}).
Alongside, ProteinChat~\citep{guo2023proteinchat} has utilized LLMs for knowledge retrieval to facilitate discussions about 3D structure databases, and concurrent to our work, Mol-Instructions \citep{fang2024molinstructions} focuses on improving LLM’s understanding of biomolecules (e.g., proteins) by fine-tuning existing LLMs. These recent efforts while introducing valuable protein-text datasets have limitations that make them unsuitable for scientific PQA, as detailed below:

\begin{enumerate}[leftmargin=*,topsep=0pt,itemsep=0pt]
    \item \textbf{ProtDescribe~\citep{xu2023protst}:} This dataset  contains textual annotations from three fields of uncurated SwissProt entries, risking bias and data leakage due to high sequence similarity and over-representation of certain protein families (Fig. \ref{uniref}). Also, its use of fields that are not directly inferable from protein sequence alone (e.g., cellular interaction) limits its utility for PQA. Conversely, it misses out on features like catalytic reactions which are often subject of scientific queries. Crucially, lacking QAs, it fails to directly support PQA assessments. Therefore, while valuable for enhancing protein representations and text-to-protein generation, as intended by the authors, this dataset is not suitable for the scientific PQA.
    \item \textbf{PDB-QA~\citep{guo2023proteinchat} :} With a limited set of 30 predefined QAs on 3D structures from the PDB database, while suitable for knowledge retrieval tasks, its focus on entry-specific details (e.g., submission date, analysis software) renders it ineffective in the context of scientific PQA.
    \item \textbf{Mol-Instructions~\citep{fang2024molinstructions}:} This dataset was released concurrent to our work. The protein section of Mol-Instructions represents a relevant set of template-based textual annotations for proteins tailored towards five downstream tasks, including question-answering. The dataset is curated from SwissProt and is debiased using a 90\% similarity threshold. This lenient threshold results in an abundance of highly related proteins, which leads to bias and in the absence of an evolution-aware splitting strategy could cause leakage across data splits.    
\end{enumerate}

Furthermore, in all previous and concurrent research authors have relied on BLEU or ROUGE metrics for evaluating the scientific accuracy of generate captions or responses. However, these metrics are not suitable for assessing accuracy \citep{Mathur2020}, therefore limiting the scientific scope of past research in PQA domain. As a result, due to the lack of relevant benchmarking strategies as well as biases and noise in existing datasets, the task of free-form zero-shot scientific enquiry of new protein sequences remains unexplored.

\begin{figure}[t]
	\begin{center}
	\includegraphics[width=1.0\textwidth]{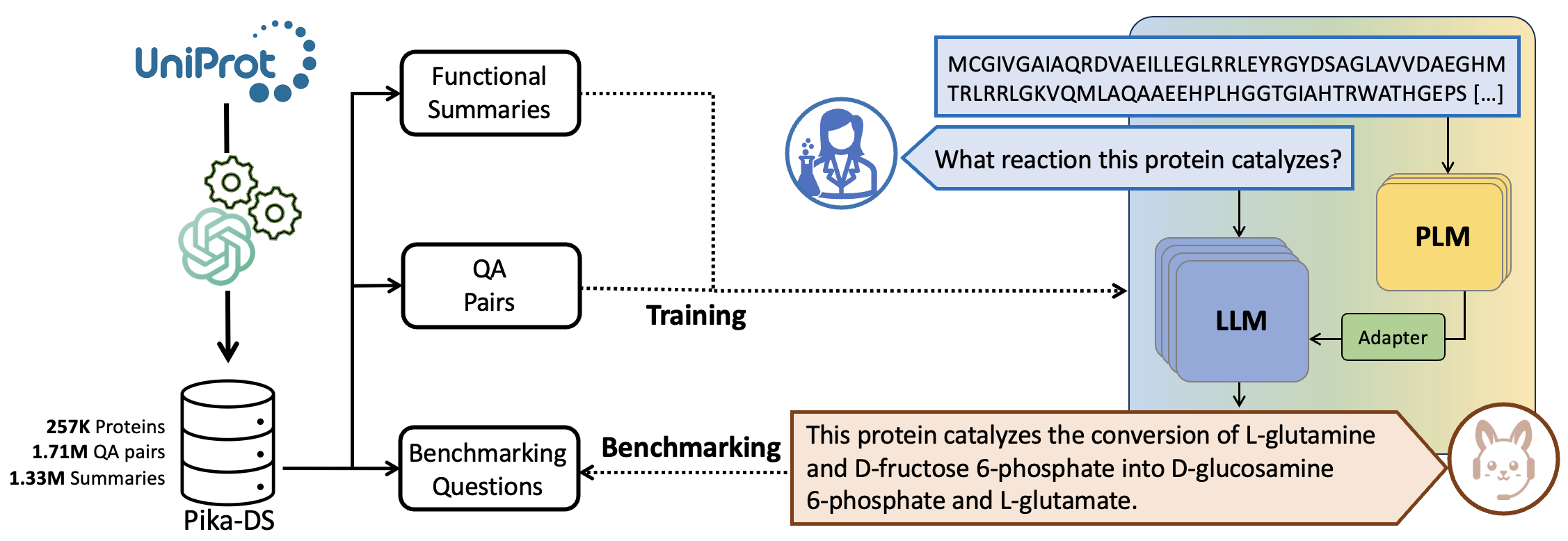}
	\end{center}
        \captionsetup{font=small, belowskip=-5pt}
	\caption{Schematic of Pika framework. Pika-DS is created from filtered SwissProt entries followed by processing using GPT3.5.}	
	\label{fig_golden_obj}
\end{figure}

\section{Pika Framework}

Here we detail our dataset, benchmarking and baseline designs for the Pika framework (Fig. \ref{fig_golden_obj}). All baselines, model architectures and benchmarking were implemented in PyTorch Lightning , and the complete codebase is accessible on \href{github.com/EMCarrami/Pika}{github.com/EMCarrami/Pika}.

\subsection{Pika Dataset}

The scientific PQA task is aimed at delivering accurate responses to free-form questions based on an unseen protein sequence. This task emerges from the need for scientific exploration of protein functions via natural language question answering. Therefore, the training and evaluation of multimodal models for the PQA task necessitates comprehensive datasets with scientific textual annotations linked to corresponding protein sequences. We deemed the following three criteria as essential for a specialized PQA dataset: 
\begin{enumerate}[leftmargin=*,topsep=0pt,partopsep=0pt,itemsep=5pt,parsep=0pt]
    \item Offers an unbiased representation of known protein sequences, mitigating frequency biases prevalent in existing databases.
    \item The dataset ensures that the information associated with each protein sequence is expertly curated, allowing for inference solely based on the protein sequence.
    \item Supports relevant benchmarking to assess model performance especially in zero-shot settings.
\end{enumerate}

Since, as discussed in section \ref{sec:LitPQA}, none of the existing datasets meet all these criteria we created Pika-DS (See sections \ref{creation}, \ref{sec:uniprot}, \ref{sec:uniref} for details), the first specialized and debiased PQA dataset, accompanied with respective biologically relevant benchmarks for model training and evaluation (summary statistics in Table \ref{tab:dataset_count} and example in Table \ref{tab:exmp}).

Briefly, we gathered all SwissProt entries from UniProt database \citep{UniProt} and extracted an expert-curated list of scientific information fields covering a wide range of relevant properties. We then debiased the sequences using a strict 50\% similarity threshold. Finally, we employed GPT3.5 API to process the information fields for each protein entry using systematically optimized prompts (Sections \ref{promptopt} and \ref{gpt3}) to create the Pika-DS's three main components: 
    \begin{itemize}[leftmargin=*]
        \item \textit{Summary:} A summary of each protein’s functional and biochemical properties, based solely on the provided information excluding the protein's name.
        \item \textit{QAs:} Several diverse QA pairs for each information field, formatted for LLM training.
        \item \textit{Metrics:} Single-word answers to a set of predefined scientific questions serving as the ground-truth for our Biochem-Lite benchmarks (Section \ref{Sec:Bench}).
    \end{itemize}

\begin{figure*}[t]
    \centering
    \includegraphics[width=1.0\textwidth]{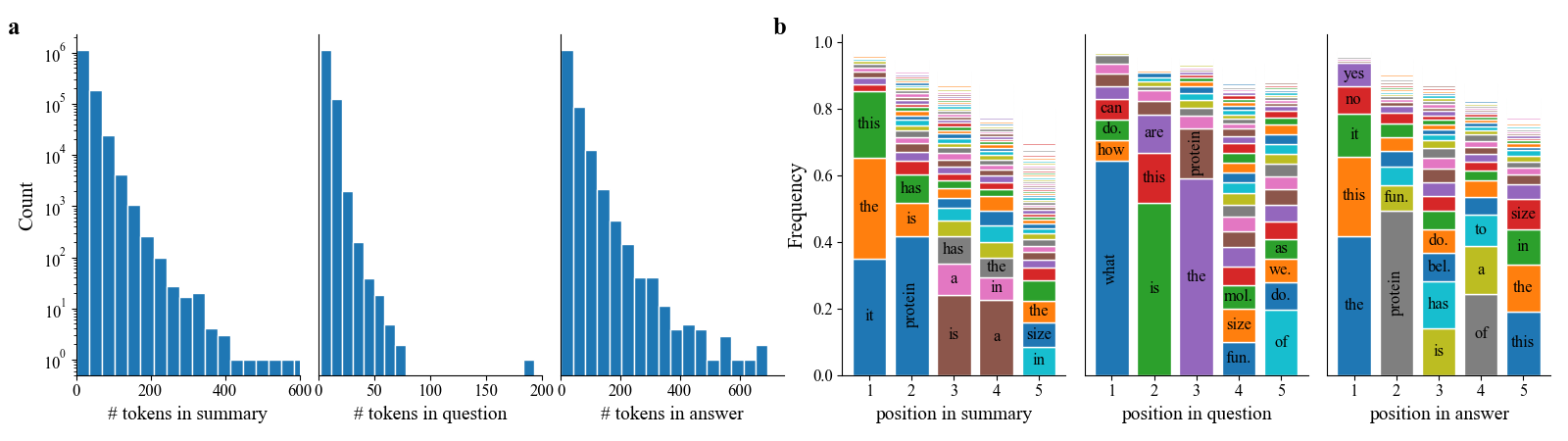}
    \captionsetup{font=small}
    \setlength{\abovecaptionskip}{-8pt}
    \setlength{\belowcaptionskip}{-5pt}
    \caption{Characteristics of PQA dataset. (a) Distribution of token counts for all examples in Pika-DS. (b) Frequency of words in each position in each section of the dataset. Long words are abbreviated (do.=does, mol.=molecule, fun.=function, we.=weight, bel.=belong).}\label{tokens}
\end{figure*}

\subsubsection{Pika-DS Quality Control}

The final Pika-DS comprises 257,167 protein sequences, selected from 185,128 UniRef50 clusters. This dataset is enriched with detailed descriptors for each sequence: a summary statement divided into sentences, multiple scientific QA pairs, and the ground-truth answers for our predefined benchmarking questions. This dataset encapsulates over 105 million protein sequence tokens, approximately 36.4 million tokens in textual summaries, and nearly 47 million tokens in QA pairs (Table \ref{tab:dataset_count}). To ensure the high quality of Pika-DS and suitability of generated textual annotations for training multimodal LLMs we assessed various aspects of our dataset.

\paragraph{Human evaluation of GPT3.5 generated annotations:} Given the potential limitations of GPT3.5 generated annotations (e.g., hallucination), we conducted a thorough human evaluation of the Pika-DS. For this, 100 randomly selected examples from Pika-DS were manually evaluated by an expert biochemist. Our analysis revealed that out of a total of 1204 summary sentences and QA pairs, 5 were incorrect (0.4\%), 21 were of poor quality (1.7\%), and 43 were irrelevant (3.6\%), resulting in over 94\% of the annotations being correct, relevant, and of high quality. Furthermore, only 5 out of 660 metric ground truths were found to be incorrect, yielding a 99.2\% accuracy in ground truth metric values. Furthermore, consultations with two expert biologists confirmed that the QA pairs generated by GPT-3.5 in Pika-DS are relevant and scientifically sound, closely matching what they would derive from the provided input fields. They also unanimously agreed that GPT-3.5 questions offer greater diversity compared to their potential queries. These results underscore the effectiveness of our prompt optimization approach for generating high-quality labels using GPT3.5 based on SwissProt information fields and is consistent with a recent study that has found LLM-generated captions can exhibit higher diversity than those created by humans, leading to enhanced training of multimodal LLMs \citep{sharifzadeh2024synth2}.

\paragraph{Token Count Analysis: } In creating the QA pairs using GPT3.5, we instructed the model to  cover a broad spectrum of queries and to produce detailed answers suitable for training other LLMs, anticipating elaborate rather than single-word answers. This was confirmed by our analysis of token counts in the summary sentences, questions, and answers of the final dataset. This revealed that while questions are typically shorter (8.5 tokens on average), the summary sentences and answers are longer and more extensive (on average 27 and 16.5 tokens). This presents an adequate number of tokens per example, providing sufficient context for the training of multimodal PQA models (Figure \ref{tokens}a).

\paragraph{Word Frequency Analysis:} To ensure the diversity of content in summary statements as well as QA pairs, we examined the frequency of words in the initial five positions of each of these categories. As anticipated, common words such as "the" and "it", in summaries, and interrogatives like "What" and "How" in questions, dominated the first three positions. Similarly, answers often began with "the" or "this" as well as "yes/no" structures, followed by "protein". However, the remaining positions demonstrated a significant lexical diversity. This observation confirmed that our dataset does not exhibit a substantial bias, offering the necessary diversity for effective model training (Figure \ref{tokens}b).

\paragraph{Protein over-representation Analysis:} We used pre-computed protein embeddings from UniProt database and visualized them using UMAP \citep{mcinnes2018umap}, highlighting the entries that belong to the top 100 largest clusters. Visual comparison of distribution of protein embeddings before and after our Uniref50-based filtering confirms a strong reduction of over-represented protein families in Pika-DS while maintaining the rich diversity of the dataset (Figure \ref{umaps}).

\subsection{Pika Benchmarks}

In this section, we elaborate on our biochemically-focused benchmarking methodologies tailored for evaluating the scientific accuracy of multimodal PQA models.

\subsubsection{Motivation \& Design Criteria}

Conventional linguistic metrics like BLEU and ROUGE, while useful in general linguistic contexts, often fall short in assessing scientific correctness and show poor correlation with human judgment \citep{Mathur2020}. As a result these metrics are inadequate for assessing the performance of multimododal PQA models. Therefore, going beyond standard linguistic evaluations, we designed a purpose-built benchmarking approach incorporating a set of predefined, biochemically-significant questions, that are specifically selected to:
\begin{enumerate}[leftmargin=*, topsep=0pt, partopsep=0pt]
    \setlength{\itemsep}{5pt}   
    \setlength{\parsep}{0pt}
    \setlength{\parskip}{0pt}
    \item Reflect the biochemical properties of proteins, ensuring that a model's accuracy in these responses is indicative of its effectiveness in broader scientific enquiries.
    \item Span a spectrum of complexity, from straightforward information extraction with minimal linguistic intricacy to advanced linguistic reasoning for identifying pertinent information. 
\end{enumerate}

In consultation with domain experts, we identified five core protein properties at distinct difficulty levels that form the basis for our scientific benchmarking questions: \textit{molecular weight} (mw), \textit{co-factor binding}, \textit{sub-cellular localization}, \textit{protein domains}, and \textit{enzymatic reaction}.

Alongside these benchmarking questions, we also require a robust strategy to assess the correctness of open-ended answers to them. Four possible assessment strategies could be imagined: (1) Exact matching of statements, (2) Keyword comparison, (3) Comparison using stronger LLMs (xLLMs), and (4) Human evaluation. While exact statement matching is useful in some research domains, it lacks the rigour required for assessing scientific accuracy. On the other hand, human evaluation for specialized domains is extremely costly at large scales. However, we can use human supervision to ensure the quality of comparisons conducted by xLLMs. As a result we selected \textit{keyword comparison} and \textit{use of xLLMs} guided by human experts for Pika framework.

\subsubsection{Two-Tiered Benchmarking System}
\label{Sec:Bench}

The use of xLLMs for response evaluation is more accurate but computationally prohibitive as compared to keyword comparison, which pragmatically balances scientific accuracy with computational efficiency. This balance of performance vs efficiency motivated us to design a two-tiered benchmarking system comprised of a light-weight benchmark employing keyword comparison for efficiency, and a rigorous xLLM-based benchmark for scientific fidelity.

{\renewcommand{\arraystretch}{1.2}
\begin{table*}[ht]
    \resizebox{0.99\textwidth}{!}{
    \begin{tabular}{l|l|l|l}
        \hline
        \textbf{Metric ID} & \textbf{Question} & \textbf{Example Label} & \textbf{Example Response} \\
        \hline
        mw MALE {$\downarrow$} & \begin{tabular}[c]{@{}l@{}}What is the molecular weight of\\ this protein?\end{tabular} & \multicolumn{1}{l|}{55808} & \begin{tabular}[c]{@{}l@{}}The molecular  weight of this protein is\\ \underline{54988} KDa.\end{tabular} \\ \cline{3-4}
        \hline
        \raisebox{-3.5pt}{exact cofactor {$\uparrow$}} & \raisebox{-3.5pt}{What is a cofactor of this protein?} & \multicolumn{1}{l|}{\raisebox{-3.5pt}{Zn(2+)}} & \raisebox{-3.5pt}{The cofactor of this protein is \underline{Zn(2+)}.} \\[7pt] \cline{3-4}
        \hline
        is\_enzyme F1 {$\uparrow$} & \begin{tabular}[c]{@{}l@{}}Can this sequence be considered\\ an enzyme? \end{tabular} & \multicolumn{1}{l|}{True} & \underline{Yes} \\ \cline{3-4}
        \hline
        location F1 {$\uparrow$} & Where is this protein located? & \multicolumn{1}{l|}{Membrane} & \begin{tabular}[c]{@{}l@{}}The sub-cellular location of this protein \\ is the cell \underline{membrane}.\end{tabular} \\ \cline{3-4}
        \hline
        \begin{tabular}[c]{@{}l@{}}binary Q\\ mean F1 \smash{\raisebox{1.2ex}{\resizebox{0.6em}{2.5ex}{$\uparrow$}}}\end{tabular} & \begin{tabular}[c]{@{}l@{}}Is this a membrane protein?\\ Is this a nuclear protein?\\ Is this a mitochondrial protein?\end{tabular} & \multicolumn{1}{l|}{\begin{tabular}[c]{@{}l@{}}True\\ False\\ False\end{tabular}} & \begin{tabular}[c]{@{}l@{}}\underline{Yes}, it is a single-pass membrane protein.\\ \underline{No}, it is localized to the cytoplasm.\\ \underline{No}\end{tabular} \\ \cline{3-4}
        \hline
    \end{tabular}}
    \captionsetup{font=small}
    \setlength{\belowcaptionskip}{-1pt} 
    \caption{Questions and example responses for Biochem-Lite (See \ref{sec:QS} for our motivation for each metric). \underline{Underlined} indicates extracted entities for score calculation. {$\uparrow$}: Higher values better. {$\downarrow$}: Lower values better.} 
    \label{tab:lw-metrics}
\end{table*}}

{\renewcommand{\arraystretch}{1.2}
\begin{table*}[ht]
    \resizebox{0.99\textwidth}{!}{
    \begin{tabular}{l|l|l|l}
    \hline
    \textbf{Metric ID} & \textbf{Question} & \textbf{Ground Truth} & \textbf{Example Response} \\
    \hline
    Reaction & \begin{tabular}[c]{@{}l@{}}What chemical reaction is \\ catalyzed by this protein?\end{tabular} & \begin{tabular}[c]{@{}l@{}}EC = 5.1.1.1,\\ L-alanine \(\rightarrow\) D-alanine\end{tabular} & \begin{tabular}[c]{@{}l@{}}This protein catalyzes the conversion of \\ L-alanine to D-alanine.\end{tabular} \\
    \hline
    Domains & \begin{tabular}[c]{@{}l@{}}What are the functional \\ domains of this protein?\end{tabular} & \begin{tabular}[c]{@{}l@{}}PLP-binding barrel, Alanine racemase, \\ Alanine racemase C-terminal domain-like, \\ pyridoxal phosphate binding\end{tabular} & \begin{tabular}[c]{@{}l@{}}The functional domains of this protein \\ include the alanine racemase domain, \\ Alanine racemase C-terminal domain-like, \\ Alanine racemase, and   PLP-binding barrel.\end{tabular} \\
    \hline
    Cofactors & \begin{tabular}[c]{@{}l@{}}What are the cofactors \\ of this protein?\end{tabular} & pyridoxal 5'-phosphate & \begin{tabular}[c]{@{}l@{}}The cofactors of this protein are \\ pyridoxal 5'-phosphate and magnesium ions.\end{tabular} \\
    \hline
    \end{tabular}}
    \captionsetup{font=small}
    \setlength{\belowcaptionskip}{-10pt}
    \caption{Questions and example responses for Biochem-ReAct. Presented in order of difficulty, with \textit{Reaction} being the most difficult question for a multimodal LLM to answer based on a protein sequence.}
    \label{tab:react-metrics}
\end{table*}}

\begin{enumerate}[leftmargin=*]
    \item \textbf{Biochem-Lite:} Pika's light-weight benchmarking involves a set of pre-defined, scientifically relevant questions with simple answers extracted for each protein using GPT3.5 during Pika-DS creation (Table \ref{tab:lw-metrics}). These questions are designed to cover a range of biochemical and functional properties (e.g. binding to cofactros or cellular localization) while evaluating multimodal PQA performance at various levels (i.e., information extraction from protein embeddings, cross-modal information processing by the LLM, or the LLM’s ability to generate relevant responses). Although the ground truth answers to these questions are single-worded, for evaluation, the questions are presented in free-form, and the model provides open-ended responses, which is processed via rule-based entity extraction and is scored using adequate metrics as detailed in Section \ref{sec:formulas}.

    \item \textbf{Biochem-ReAct:} While light-weight benchmarking questions offer a general overview of model's performance in various scientific aspects, they do not represent real-world scenarios where the scientific fidelity of open-ended responses is vital. Therefore, we also selected three biochemical questions at three distinct difficulty levels surrounding \textit{Reaction \& Activity} of the proteins that can be subjects of scientific enquiry into novel protein sequences (Table \ref{tab:react-metrics}).
    The complexity of the responses to these questions necessitates assessment using advanced LLMs such as GPT4. To ensure the high reliability of these assessments we performed an iterative prompt optimization with human feedback following the approach described in Section \ref{promptopt}. 
\end{enumerate}

\vspace{-5pt}
Overall, our two-tiered benchmarking system establishes a balanced and robust framework for evaluating multimodal PQA models, ensuring both scientific rigor and computational feasibility.

\subsection{Pika Baselines}
As discussed in Section \ref{sec:LitPQA}, previous PQA studies have either been limited to knowledge retrieval or have ignored the evolutionary context of protein sequences. Combined with the use of unsuitable metrics such as BLEU, these studies do not provide relevant benchmarking opportunities for scientific PQA. Furthermore, unlike natural multimodal LLMs (e.g., Vision Question Answering) where human-baselines are possible, in PQA it is not possible to determine the upper-bound of performance by comparison to expert evaluations, because humans are unable to extract any information from protein sequences. Therefore, in the absence of relevant baselines or human evaluation for PQA, we introduce two multimodal PQA-LLMs and define various lower- and upper-bound baselines. 

\begin{figure*}[ht]
	\begin{center}
		\includegraphics[width=1.0\textwidth]{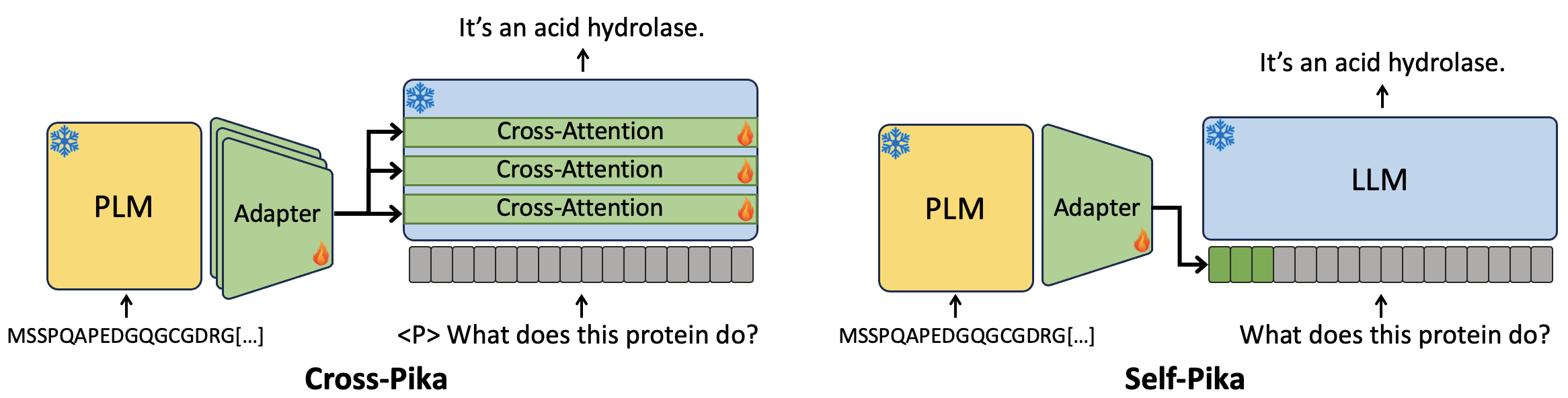}
	\end{center}
         \captionsetup{font=small}
         \setlength{\belowcaptionskip}{-5pt}
         \setlength{\abovecaptionskip}{5pt}
	\caption{Schematic representation of Cross- and Self-Pika architectures for the scientific PQA task. PLM = Protein Language Model (protein sequence encoder), LLM = Large Language Model. Only the adapter and cross-attention modules (both in green) are trained.}\label{arch}
\end{figure*}

\subsubsection{Pika Models}
We propose two robust multimodal architectures for PQA tasks, drawing inspiration from successful Vision Language Model (VLM) strategies, replacing the vision-encoder with a protein language model (PLM) (Fig. \ref{arch}). 
\begin{enumerate}[leftmargin=*]
    \item \textbf{Cross-Pika} is inspired from \textit{Flamingo}~\citep{tsimpoukelli2021multimodal} and \textit{Prismer}~\citep{liu2023prismer}. It uses multiple independent learnable adapters, each creating distinct protein latent embeddings for each transformer layer of the LLM. These embeddings are injected into the LLM using a gated cross-attention mechanism before each native self-attention layer. At the cost of increased complexity, this design allows for a nuanced and layer-specific modulation of the LLM, potentially enabling it to process complex protein-related information more effectively, which could aid the model in answering biochemically intricate PQA queries.
    \item \textbf{Self-Pika} is based on the architecture proposed in \textit{Frozen}~\citep{tsimpoukelli2021multimodal} and uses a single learnable adapter to transform protein sequence embeddings into latent embeddings compatible with language token embeddings. The transformed embeddings, concatenated at the beginning of the LLM's initial token embeddings, allow the protein latent embeddings to influence the LLM’s response through its internal self-attention mechanisms. This approach effectively creates soft-prompts conditioned on the input protein sequence. This architecture simplifies the integration process, reducing computational overhead while  allowing for the incorporation of essential protein characteristics into the LLM's response.
\end{enumerate}
Considering the diversity in protein lengths, we opted for the Perceiver architecture \citep{jaegle2021perceiver} as the learnable adapter for both models. This choice standardizes the transformation of protein embeddings into a consistent number of latent embeddings for seamless cross-modality information transfer. In both architectures we keep the pre-trained LLM and PLM frozen, only incorporating trainable "adapter" modules that extract relevant latent representations from PLM's output (protein sequence embeddings) and integrate them into the LLM's transformer architecture, facilitating seamless transfer of information across modalities. In this work we use ESM2 \citep{lin2022language} as the pre-trained PLM and Phi-2 \citep{huggingface2023phi2} as the pre-trained LLM (In some experiments GPT2 \citep{radford2019language} was used, where indicated).

\subsubsection{Lower-bound Baselines} \label{sec:lowbase}
\begin{itemize}[leftmargin=*, topsep=0pt, partopsep=2pt]
    \item \textbf{Random Baseline:} Selects a random answer for each question from the pool of relevant answers.
    \item \textbf{LLM Only:} Pre-trained Phi-2 model to determine LLM's base response to Biochem-Lite questions. 
    \item \textbf{Pika w/o PLM:} Mirrors Self-Pika's architecture and training, substituting the PLM with simple token embeddings of the protein sequences. This effectively reduces the information to amino acid content only, ignoring the context and order of the sequence. 
\end{itemize}

\subsubsection{Upper-bound Baselines}
\begin{itemize}[leftmargin=*, topsep=0pt, partopsep=2pt]
    \item \textbf{MLP:} An MLP on protein sequence embeddings of ESM2 (Details in Section \ref{sec:mlp}). It defines the upper limit for the information content of ESM2 embeddings, with any performance beyond this by a model potentially indicating generalization due to the knowledge encapsulated in the LLM.
    \item \textbf{BLAST:} Considering the evolutionary context of protein sequences, for any queried sequence, we identify the closest related protein sequence in the training data using BLAST \citep{Camacho2009BLAST} and return the respective information as the prediction. This is a known strong upper-bound baseline, as homology is a strong predictor of protein's properties.
\end{itemize}

\section{Experiments \& Results}

In this section we share our key experimental results. Training details can be found in Sections \ref{sec:training} and \ref{sec:hyperopt}.

\begin{figure*}[t]
    \centering
        \includegraphics[width=1.0\textwidth]{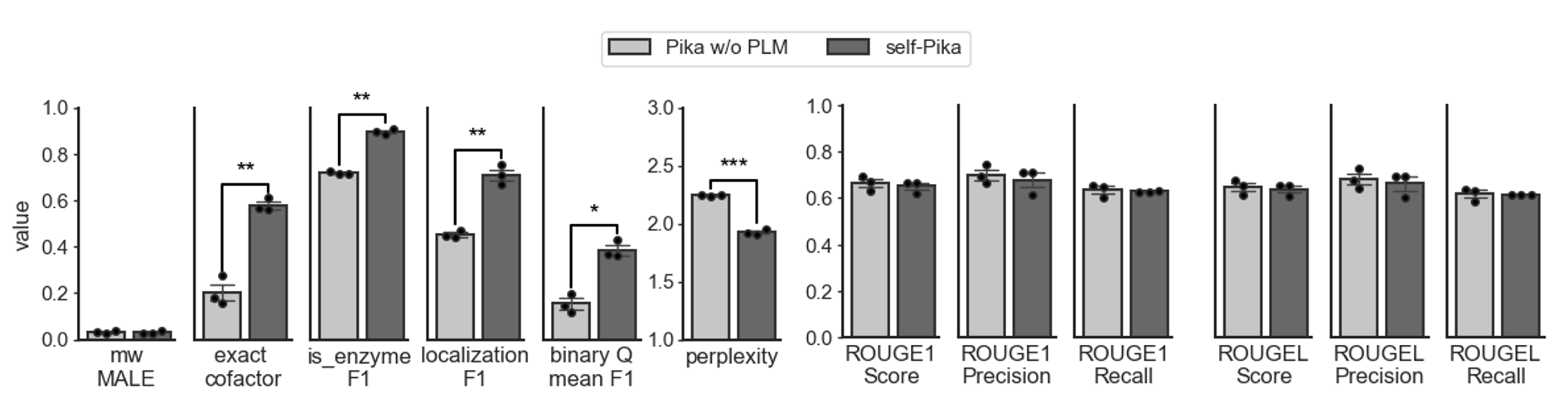}
         \captionsetup{font=small}
         \setlength{\abovecaptionskip}{-8pt}
         \setlength{\belowcaptionskip}{-10pt}
    \caption{Evaluating the effectiveness of Biochem-Lite vs traditional linguistic metrics for scientific accuracy of PQA. Statistical significance is determined through a one-tailed paired t-test across three randomly seeded data subsets and model training (significance guide: \(^{*}p < 0.05\), \(^{**}p < 0.01\), \(^{***}p < 0.001\)). 
    }\label{stats}
\end{figure*}

\subsection{Validation of Biochem-Lite} 

First, we assessed the reliability of Biochem-Lite metrics in assessing PQA models performance. To We compared a simple Pika model (Self-Pika with ESM2 + GPT2) against its truncated version, Pika w/o PLM (Section \ref{sec:lowbase}),  where ESM2 model was replaced with simple protein token embeddings. Considering that the removal of ESM2 should eliminate all functional information, leaving behind only the amino acid content, we expect that the truncated model must perform significantly worse than the original model. Therefore, metrics that fail to show a significant difference between these two would be unsuitable for assessing PQA models. Our results indicate that Biochem-Lite metrics are significantly better in Self-Pika vs Pika w/o PLM, while all ROUGE metrics fail to highlight any differences (Fig. \ref{stats}). The behaviour of the \textit{mw MALE} metric is expected as the size of a protein only depends on its amino acid content. This finding confirms the inadequacy of traditional linguistic metrics for scientific PQA task, demonstrating the utility and importance of our Biochem-Lite metrics.

\subsection{Zero-shot PQA:}
We define zero-shot PQA as answering questions about unseen protein sequences. Considering the debiased nature of Pika-DS, this can easily be achieved by splitting the data based on UniRef50 ID of each sequence, thereby ensuring that the validation/test sequences have no more than 50\% sequence similarity with any of the training sequences. While this ensures the \textit{unseen sequence} criteria, when considering the evolutionary context of proteins, the 50\% similarity threshold is a lenient cutoff. This is because homologous proteins may exhibit as low as 15\% sequence similarity \citep{homology}. As a result, on UniRef50-based splits, model performance will be a combination of generalization based on the sequence and model's ability to identify distant evolutionary relations to training data. While such behaviour may be desired for some PQA applications, to extend the scope, we created a more stringent splitting strategy that focuses the evaluation of model's generalization primarily on the basis of proteins sequences without the impact of evolutionary relationships. To achieve this we minimized the the evolutionary connection of validation/test sequences to the training data. This was achieved by first grouping sequences into evolutionary groups (EvoGroups) (See Section \ref{sec:evogroups} for details) followed by splitting the data based on their EvoGrop assignments, ensuring that sequences in validation and test do not have closely related evolutionary counterparts in the training data. Naturally, we expect that the BLAST baseline should show a strong performance on UniRef50-based splits, while its performance should diminish on EvoGroup-based splits.

{\renewcommand{\arraystretch}{1.3}
\begin{table*}[ht]
    \small
    \centering
    \resizebox{0.99\textwidth}{!}{
    \begin{tabular}{ll|ccccc|ccc}
        \hline
        \multicolumn{2}{c|}{\multirow{2}{*}{\raisebox{-20pt}{\textbf{Baseline or Model}}}} & \multicolumn{5}{c|}{\textbf{Biochem-Lite}} & \multicolumn{3}{c}{\textbf{Biochem-ReAct}} \\ 
        \cline{3-10}
         & & \begin{tabular}[c]{@{}c@{}}mw\\ MALE\end{tabular} & \begin{tabular}[c]{@{}c@{}}exact\\ cofactor\end{tabular} & \begin{tabular}[c]{@{}c@{}}is\_enzyme\\ F1\end{tabular} & \begin{tabular}[c]{@{}c@{}}location\\ F1\end{tabular} & \begin{tabular}[c]{@{}c@{}}binary\\ mF1\end{tabular} & Reaction & Domains & Cofactor \\
         \hline
        & Random & 0.35 (0.01) & 0.13 (0.02) & 0.49 (0.00) & 0.34 (0.02) & - & - & - & -\\
        Lower-bound & Phi-2 only & 4.53 (0.01) & 0.03 (0.01) & 0.34 (0.00) & 0.00 (0.00) & \underline{0.31} (0.00) & - & - & - \\
        & Pika w/o PLM & \textbf{0.02} (0.01) & 0.21 (0.01) & 0.7 (0.01) & 0.47 (0.01) & 0.21 (0.08) & 0.01 & 0.32 & 0.82 \\
        \hline
        \multirow{2}{*}{Upper-bound} & MLP & 0.07 (0.01) & - & 0.86 (0.01) & \underline{0.89} (0.03) & - & - & - & - \\
        & BLAST & - & \textbf{0.71} (NA) & \underline{0.88} (NA) & \textbf{0.93} (NA) & - & \textbf{0.94} & \textbf{0.89} & \textbf{0.99} \\
        \hline
        \multirow{2}{*}{Pika Models} & Cross-Pika & \textbf{0.02} (0.00) & 0.47 (0.15) & \textbf{0.89} (0.02) & 0.77 (0.01) & 0.13 (0.02) & 0.36 & 0.60 & 0.94 \\
        & Self-Pika & \underline{0.04} (0.02) & \underline{0.54} (0.04) & \textbf{0.89} (0.01) & 0.76 (0.02) & \textbf{0.44} (0.04) & \underline{0.58} & \underline{0.79} & \underline{0.97} \\
         \hline
    \end{tabular}}
    \setlength{\belowcaptionskip}{-8pt}
    \captionsetup{justification=raggedright,singlelinecheck=false, font=small}
    \caption{Performance of Pika models on Uniref50 splits. \textbf{Bold} values indicate the best score, \underline{underline} indicates second best score. Values in ( ) indicate standard deviation. Biochem-Lite results on val set and from three different seeded training. Biochem-ReAct results on test set and a single seed.}
    \label{tab:res_table}
\end{table*}}

\subsubsection{Performance on UniRef50 Splits}

 Table \ref{tab:res_table} summarizes the benchmarking results of Pika models on a random UniRef50-based split in comparison with various baselines. It is evident that both Pika architectures outperform all lower-bound baselines, indicating successful cross-modality information transfer to the LLM. Notably, our top model surpasses the upper-bound MLP baseline on key Biochem-Lite benchmarking questions, underscoring the potential for generalization via the LLM's knowledge. However, as expected, the strong BLAST baseline outperforms all models in most Biochem-Lite and all Biochem-ReAct metrics. Nevertheless, these results indicate that our Pika models based on the modestly sized Phi-2 LLM, are able to comprehend distant evolutionary relations form the context of sequence embeddings.

{\renewcommand{\arraystretch}{1.2}
\begin{table*}[ht]
    \small
    \centering
    \resizebox{0.75\textwidth}{!}{
    \begin{tabular}{l|ccc|ccc}
        \hline
         \multirow{2}{*}{\raisebox{-10pt}{\textbf{Baseline or Model}}} & \multicolumn{3}{c|}{\textbf{Biochem-Lite}} & \multicolumn{3}{c}{\textbf{Biochem-ReAct}} \\ 
        \cline{2-7}
         & \begin{tabular}[c]{@{}c@{}}exact\\ cofactor\end{tabular} & \begin{tabular}[c]{@{}c@{}}is\_enzyme\\ F1\end{tabular} & \begin{tabular}[c]{@{}c@{}}location\\ F1\end{tabular}  & Reaction & Domains & Cofactor \\
         \hline
        BLAST Baseline  & 0.21 & 0.72 & 0.51 & \textbf{0.52} & 0.49 & \textbf{0.94} \\
        Self-Pika Model  & \textbf{0.57} & \textbf{0.85} & \textbf{0.69} & 0.09 & \textbf{0.54} & 0.87 \\
         \hline
    \end{tabular}}
    \setlength{\belowcaptionskip}{-8pt}
    \captionsetup{justification=raggedright,singlelinecheck=false, font=small}
    \caption{Performance of Pika model on EvoGroup splits. \textbf{Bold} values indicate the best score. Biochem-Lite results on val set and Biochem-ReAct results on test set both with a single seed.}
    \label{tab:res_table2}
\end{table*}}

\subsubsection{Performance on EvoGroup Splits}

To assess generalization by Pika models, in isolation from the effects of evolutionary relationship of proteins, we performed Self-Pika training on EvoGroup splits and compared the results with BLAST baseline on these splits (Table \ref{tab:res_table2}).  As expected, our results show a significant reduction in performance of BLAST baseline, specially in more complex metrics such as \textit{Domains} and \textit{Reaction}, and to a lower extent in simpler metrics such as \textit{Cofactor}. This is because, cofactor binding, for instance, relies on small protein motifs that are more conserved across distant evolutionary relations. Crucially, Self-Pika, except on the most complex \textit{Reaction }metric, retained the majority of its performance on EvoGroup splits, allowing it to surpass the performance of BLAST baseline in most metrics. These observations indicate that multimodal PQA models are capable of inferring functional properties both based on the sequence of proteins as well as using distant evolutionary relations.   

\subsubsection{Correlation of Biochem Metrics} 

The Self-Pika model demonstrated strong performance across all three Biochem-ReAct questions on the UniRef split, exceeding the \textit{without PLM baseline} by 57\%, 47\%, and 15\% for correctly identifying \textit{Reactions}, \textit{Domains}, and \textit{Cofactors}, respectively, for previously unseen protein sequences. This prompted us to study the correlation of Biochem-Lite metrics to Biochem-ReAct metrics. Comparing Biochem-Lite results against Biochem-ReAct scores for 12 high-performing checkpoints with varied configurations revealed that while traditional linguistic metrics, including perplexity, fall short in predicting the real-world efficacy of PQA models, two Biochem-Lite questions, "exact cofactor recall" and "binary Q mean F1", emerged as strong indicators of multimodal PQA models’ real-world performance based on Biochem-ReAct metrics (Fig. \ref{corrb}).

\subsubsection{PQA Learning without Questions}

To understand the significance of QA pairs in the training set, we conducted training under two stringent conditions. The first involved using only summary sentences as labels for proteins without any QAs. The second condition included summary sentences and a single Control Question, "Is this a real protein?". During training, we randomly shuffled the tokens within the sequence of half the proteins, setting the expected answer to this question as No (for summary labels all protein sequences remained unchanged). This aims to train the model to understand the task of question answering, with summaries providing the scientific context for learning. Remarkably, introducing the Control Question, even when no other questions were present during training, significantly enhanced performance (Table \ref{tab:zeroQ}). Additionally, we observed that the performance of the model on the Control Question follows a similar pattern as other metrics (Table \ref{tab:controlQT}). These observation suggest that the current bottleneck in PQA performance is likely the LLM and extending this work to larger LLMs could further improve performance and generalization capabilities.

\captionsetup[table]{belowskip=0pt}
{\renewcommand{\arraystretch}{1.2}
\begin{table*}[ht]
    \centering
    \resizebox{0.99\textwidth}{!}{
    \begin{tabular}{lccc|ccc|ccc}
        \hline
        \multicolumn{4}{c|}{Training Mode} & \multicolumn{3}{c|}{Biochem-Lite} & \multicolumn{3}{c}{Biochem-ReAct} \\
        \hline
         Model & S &  Ctrl Q &  QA & \begin{tabular}[c]{@{}c@{}}mw\\ MALE\end{tabular} & \begin{tabular}[c]{@{}c@{}}exact\\ cofactor\end{tabular} & \begin{tabular}[c]{@{}c@{}}binary Q\\ mean F1\end{tabular} & Reaction & Domains & Cofactor \\
         \hline
         self-Pika (S) & \checkmark & \ding{55} & \ding{55} & 4.54 & 0.01 &  0.00 & - & - & - \\
         self-Pika (S+C) & \checkmark & \checkmark & \ding{55} & \underline{2.30} & \underline{0.32} &  \underline{0.42} & 0.22 & 0.70 & 0.95 \\
         self-Pika (Q) &\ding{55} & \checkmark & \checkmark & \textbf{0.04} & 0.54  & 0.44 & \textbf{0.58} & \textbf{0.79} & \textbf{0.97}  \\
         self-Pika (Q+S) &\checkmark & \checkmark & \checkmark & \textbf{0.04} & \textbf{0.57} & \underline{\textbf{0.57}} & 0.53 & 0.75 & 0.96 \\
        \hline
    \end{tabular}}
    \captionsetup{font=small}
    \setlength{\belowcaptionskip}{0pt}
    \caption{Perfomance of Self-Pika in the absence of QAs during training. \textbf{Bold} values indicate best score. Standard deviation for all scores $<0.05$ except for those indicated by \underline{underline}. S = Summary, Ctrl Q = Control Question, QA = QA pairs.}
    \label{tab:zeroQ}
\end{table*}}

\subsubsection{PQA Learning with Novel Proteins} 

In training multimodal Pika models, we ensured that proteins similar to those in the training dataset were excluded from the test and validation sets, effectively creating a zero-shot scenario for our benchmarking. Given that the ESM2 model was pre-trained on a vast dataset of proteins, there was a potential concern that its embeddings might be influenced by previously encountered sequences. To mitigate this, we utilized the evaluation split reported for ESM2's pre-training, categorizing our benchmarking results into proteins seen and unseen by ESM2. The analysis revealed no significant performance difference between these conditions, affirming that Pika's zero-shot capabilities are not compromised by the prior knowledge embedded in ESM2. This finding underscores the robustness of Pika models in genuine zero-shot PQA tasks, independent of ESM2's pretraining exposure.

\subsubsection{Ablation Studies} 

To identify key contributing factors, we compared model performance across various dimensions. Most notably, reducing the size of the LLM from Phi-2, which has 2.8 billion parameters, to GPT-2 Medium, with 355 million parameters, resulted in a 20\% decrease in the key "exact cofactor" metric (Table \ref{tab:abel}). However, the size of the ESM2 model did not seem to have a significant impact, with performance remaining largely similar between ESM2-S, ESM2-M, and ESM2-L (8M, 65M and 350M parameters, respectively). This observation could be attributed to the richness of representations in the Pre-trained PLMs, potentially highlighting a bottleneck in the LLM or the quality and quantity of the data (Table \ref{tab:abel}). Concordantly, the comparison of results from the zero-shot experiment also confirms that with a lower volume of data, a tailor-made dataset could yield better results. However, as the dataset size increases—a scenario analogous to the image domain—this factor becomes less crucial. These findings suggest that while LLM capacity significantly influences PQA model performance, the sophistication of sequence embeddings provided by ESMs reaches a point of diminishing returns, underscoring the importance of focusing on LLM enhancements and data quality for future improvements.

{\renewcommand{\arraystretch}{1.1}
\begin{table*}[ht]
    \centering
    \resizebox{0.99\textwidth}{!}{
    \begin{tabular}{llcccccccc}
        \hline
         LLM & PLM & \begin{tabular}[c]{@{}c@{}}mw\\ MALE\end{tabular} & \begin{tabular}[c]{@{}c@{}}exact\\ cofactor\end{tabular} & \begin{tabular}[c]{@{}c@{}}is\_enzyme\\ F1\end{tabular} & \begin{tabular}[c]{@{}c@{}}location\\ F1\end{tabular} & \begin{tabular}[c]{@{}c@{}}binary Q\\ mean F1\end{tabular} & perplexity \\
         \hline
         & ESM2-S & \underline{0.14} & 0.22 & 0.84 & 0.65 & \underline{0.26} & 3.26 \\
        GPT2-M & ESM2-M & 0.06 &  \underline{0.26} & 0.86 & 0.64 & 0.33 & 3.12 \\
         & ESM2-L & 0.07 &  \underline{0.32} & 0.86 & 0.70 & \underline{0.23} & 3.05 \\
         \hline
         & ESM2-S & \textbf{0.02} & 0.51 & 0.86 & 0.71 & \underline{0.33} & 1.99 \\
        Phi-2 & ESM2-M & \textbf{0.02} & \underline{\textbf{0.56}} & 0.88 & 0.75 &  \underline{\textbf{0.46}}  &  \underline{\textbf{1.95}} \\
         & ESM2-L & 0.04 & 0.54 & \textbf{0.89} & \textbf{0.76} & 0.44 &   1.98 \\
         \hline
    \end{tabular}}
     \captionsetup{font=small}
     \setlength{\belowcaptionskip}{0pt}
    \caption{Effect of LLM and PLM size on Self-Pika models' performance. Standard deviation for all scores $<0.05$ except for those indicated by \underline{underline}.}
    \label{tab:abel}
\end{table*}}

\section{Conclusion \& Impact}
We've established a pioneering framework for zero-shot PQA, presenting a significant step forward in the application of LLMs for scientific enquiry. The introduction of our specialized datasets and biologically relevant benchmarks underpins future explorations at the intersection of computational biology and artificial intelligence. Key insights from our zero-shot evaluations and ablation studies highlight the crucial role of LLMs in multimodal PQA performance, while underscoring the importance of considering evolutionary relation in assessing performance. 
Since, for Pika-DS we employed GPT3.5 to create synthetic annotations, given the limitations of LLMs in generating large-scale synthetic datasets, we endeavored to minimize the inclusion of harmful content in Pika-DS through prompt optimization and manual evaluation. Nonetheless, due to the dataset's extensive size, there's a slight possibility that unintended harmful content might still be present. Our Pika-based pre-trained models are derived from the publicly accessible and unmoderated Phi-2 LLM. Thus, all cautions, restrictions, and notes associated with \href{https://huggingface.co/microsoft/phi-2}{Phi-2} \citep{huggingface2023phi2} are applicable to our models. 
Looking ahead, leveraging larger, more diverse LLMs may offer substantial gains in model generalization, driving us towards the goal of automating accurate scientific enquiry into proteins.

\section{Reproducibility \& Accessibility}

All code and data used for creation of data, model training and baselines are publicly accessible on \url{www.github.com/EMCarrami/Pika}.

The Pika framework is specifically designed for question answering related to protein sequences. With scientists having identified nearly 0.25 billion protein sequences, and functional annotations available for fewer than a million, our framework offers significant potential for research into these largely unexplored proteins. While our efforts are directed towards scientific research, we recognize the potential risk of misuse by individuals aiming to create or identify harmful substances. We strictly prohibit using our dataset and framework for any illegal activities or actions that could harm individuals or society.


\begin{ack}
We would like to thank Dr Aida Nematzadeh for critical advice and Maryam Karaminejad Ranjbar for graphic design. We would also like to sincerely thank Relation Therapeutics for providing us with the necessary compute for training the final version of Pika. 
This work was primarily supported by personal funding and the authors declare no conflict of interest. 
\end{ack}

\bibliography{pika}
\bibliographystyle{icml2024}
\medskip

\newpage
\appendix

\section{Supplementary Methods}

\renewcommand{\thefigure}{\thesection.\arabic{figure}}
\setcounter{figure}{0} 
\renewcommand{\thetable}{\thesection.\arabic{table}}
\setcounter{table}{0} 

\subsection{Creation of Pika-DS}\label{creation}

Pika-DS was prepared in three phases:

\begin{enumerate}[leftmargin=*,topsep=0pt, partopsep=0pt]
    \setlength{\itemsep}{5pt}
    \setlength{\parsep}{2pt}
    \setlength{\parskip}{2pt}
    \item We utilized the SwissProt database as a foundational resource. SwissProt, the reviewed section of UniProt, features approximately 570,000 detailed annotations of protein sequences spanning all domains of life and viruses. For each entry, we extracted an expert-curated list of scientific information covering a wide range of subjects including evolutionary, biochemical and functional properties (Section \ref{sec:uniprot}). Removing proteins shorter than 30 amino acids, we obtained  3.7 million information fields for over 548,000 protein sequences.
    
    \item UniProt also offers similarity-based clustering of its entries at various sequence similarity thresholds of 100\%, 90\%, and 50\%, known as UniRef clusters. For instance, protein sequences that share a minimum of 50\% pairwise sequence similarity will be assigned the same UniRef50 cluster. Recognizing a strong bias towards more commonly studied protein families in SwissProt (Figure \ref{uniref}), we limited Pika-DS to a maximum of two most informative sequences per UniRef50 cluster using a custom algorithm (Section \ref{sec:uniref}). This resulted in a debiased set of over 257,000 protein sequences and 1.17 million information fields.
    
    \item Lastly, we used GPT3.5 API to process the information fields for each protein entry using systematically optimized prompts (Sections \ref{promptopt} and \ref{gpt3}) to create the Pika-DS's three main components (example in Table \ref{tab:exmp}): 
    \begin{itemize}[leftmargin=*]
        \item \textit{Summary:} A summary of each protein’s functional and biochemical properties, based solely on the provided information excluding the protein's name.
        \item \textit{QAs:} Several diverse QA pairs for each information field, formatted for LLM training.
        \item \textit{Metrics:} Single-word answers to a set of predefined scientific questions serving as the ground-truth for our Biochem-Lite benchmarks (Section \ref{Sec:Bench}).
    \end{itemize}

\end{enumerate}

\subsection{Data collection from UniProt}\label{sec:uniprot}  We retrieved XML files of all SwissProt entries longer than 30 amino acids, with the cut-off date of 14-08-2023, using UniProt's API. Since these entries contained extensive extraneous information, like author names, submission dates, etc., we employed a rule-based pre-processing approach to extract fields relevant to each protein's functional and biochemical characteristics. This selection, guided by an expert biochemist, included \textit{sequence} (molecular mass and length), \textit{organism} (top three taxonomic levels), \textit{catalytic activity} (including EC number), \textit{biophysicochemical properties} (pH and temperature dependence), \textit{cofactor}, \textit{subunit} (excluding fields containing “interact”, “associate”, or “complex”), \textit{subcellular location} (excluding isoforms), and functional domains: \textit{GO} (only molecular function, omitting biological process and cellular component), \textit{Gene3D}, and \textit{SUPFAM}.

\subsection{Debiasing SwissProt Entries Using UniRef50 Clusters}\label{sec:uniref}
To mitigate bias in SwissProt entries, we employed UniRef50 clusters, sourced from the UniProt FTP's idmapping file. First, we merged clusters representing isoforms with their corresponding main protein clusters to consolidate isoform-driven redundancy. Specifically, if the isoform of a protein was a member of another Uniref50 cluster, we merged the isoform cluster into the main protein's cluster. This step aimed to tighten clustering criteria, avoiding oversampling due to isoforms. Next, within each merged cluster, sequences shorter than 25\% of their cluster's median length were excluded to ensure a focus on sufficiently representative sequences. This filtering is due the fact that some Uniprot sequences have incomplete sequences. Finally, following the methodology outlined by Jiang et al. \citep{jiang2022more}, we calculated gzip information for each entry. Initially, the entry with the highest gzip information was identified. Subsequently, we assessed the additional gzip information provided by each remaining entry relative to this top candidate. The entry offering the highest additional gzip information was selected for inclusion. Throughout this selection process, cluster representatives were given priority in the event of a tie, ensuring the most informative representatives were chosen.
This debiasing process was designed to refine the Pika-DS by leveraging UniRef50 clusters, enhancing the dataset's quality and representativeness for downstream applications.

\subsection{GPT3.5 Prompt Optimization Strategy}\label{promptopt}
To ensure the high quality of GPT3.5 generated information and QA pairs, we performed a systematic prompt optimization focusing on GPT3.5's adherence to using only the given data, summarizing complex biological reactions accurately, and avoiding speculative or ambiguous language. This was performed in an iterative process where at each iteration, GPT3.5 was provided with the prompt and all the extracted information for 50 randomly selected protein sequences. Next, 100 outputs from each of the summary statements, QA pairs and evaluation metrics were evaluated against the input information of the respective proteins by an expert biochemist and the number of incorrect, poor quality and irrelevant (correct but unrelated to the protein's function) statements or QA pairs were noted. At each step, the prompt was modified to address the most problematic issues, with a target of no more than one incorrect, two poor quality and two irrelevant values in each category (optimised prompt in Section \ref{gpt3}).

\subsection{GPT3.5 Prompts Used for Creation of Pika-DS}\label{gpt3}

\paragraph{Summarising and QAs: } 

The following instructions were used to create summary statements and QA annotations based on information fields collected from SwissProt:

You will receive details about a specific protein.
Perform the following tasks and print each result in a new line:

1) Provide a factual summary, without using the protein's name with a maximum of 500 words.
Your summary must accurately and scientifically describe the functional, biochemical and structural properties of
this protein based only on the provided information.
Ensure that the summary follows a natural and scientific flow, starting with general information such as structure,
localization and taxonomy before detailing functional and biochemical properties.
Ensure that all key points are covered and DON'T provide any extra information than what is stated in the input.

2) For each type of information provided, create a question-and-answer pair to elucidate an aspect of
the protein's functionality or biochemical properties without using the protein's name.
Phrase your questions and answers such that they will be suitable for training a language model.
DON'T enumerate or label the questions and print each question and its answer pair in the same line.

- For all tasks if the input contains large group of properties only provide the canonical and crucial information
rather than enumerating every single entry.
- Where applicable, summarise enzymatic reactions into one or two of the generic classes of the activities.
- DON'T use any of your knowledge to add additional context or information. DON'T add any speculative or unwarranted
information that is not specifically provided.
- AVOID using generic phrases or embellished terms such as 'highly', 'several', 'diverse' and 'various'.
- Exactly follow the output format provided below to ensure consistency.
Final output format:
summary: [YOUR SUMMARY]
QA pairs:
1) [YOUR QUESTION] [YOUR ANSWER]
2) [YOUR QUESTION] [YOUR ANSWER]
[...]
n) [YOUR QUESTION] [YOUR ANSWER]

\paragraph{Biochem-Lite Ground truth: } 

The following instructions were used to collect ground-truth values for Biochem-Lite questions based on summary statements and information fields:

You will receive details about a specific protein.
Provide a single word answer to the following questions. Print each question and your answer in the same new line.
If the question does not apply to the protein, ignore the question.
1) Is this protein localized to the cell membrane?
2) Is this a membrane protein?
3) Is this protein localized to nucleus?
4) Is this protein localized to mitochondria?
5) Does this protein bind to DNA?
6) Does this protein bind to RNA?
7) Is this protein an enzyme?
8) What are co-factors of this protein as a comma separated list?

- Exactly follow the output format provided below to ensure consistency.
Final output format:
1) [MY QUESTION] [SINGLE WORD ANSWER e.g. YES/NO/UNKNOWN]
2) [MY QUESTION] [SINGLE WORD ANSWER e.g. YES/NO/UNKNOWN]
[...]
7) [MY QUESTION] [SINGLE WORD ANSWER e.g. YES/NO/UNKNOWN]
8) [MY QUESTION] [Comma separated list of co-factors e.g. FAD,FMN/UNKNOWN]

\subsection{Biochem-Lite Metric Questions and Motivation}\label{sec:QS}

\paragraph{mw MALE}
\begin{itemize}
\setlength{\itemsep}{1pt}
\setlength{\parskip}{1pt}
    \item \textbf{Question}: What is the molecular weight of this protein?
    \item \textbf{Score:} Mean Absolute Error of log values
    \item \textbf{Presence in training:} in exact form
    \item \textbf{Motivation:} Proteins are large polymers of amino acids. The molecular weight of a protein depends not only on the number of amino acids but also on the types of amino acids in the sequence.
\end{itemize}

\paragraph{Cofactor recall}
\begin{itemize}
\setlength{\itemsep}{1pt}
\setlength{\parskip}{1pt}
    \item \textbf{Question}: What is a cofactor of this protein?
    \item \textbf{Score}: If exact match, score=1; else 0 (average across examples)
    \item \textbf{Presence in training:} in exact form
    \item \textbf{Motivation}: Many proteins require other small molecular entities to function. These could be ions, or small molecules, or a combination of these, such as a Heme-Fe. Proteins often bind to their cofactors via motifs, which are short 3D structures formed due to the presence of specific sequences. Identifying the cofactor of a protein requires the extraction and analysis of these sequence motifs from the sequence embeddings.
\end{itemize}

\paragraph{is\_enzyme F1}
\begin{itemize}
\setlength{\itemsep}{1pt}
\setlength{\parskip}{1pt}
    \item \textbf{Question}: Can this sequence be considered an enzyme?
    \item \textbf{Score}: F1 score
    \item \textbf{Presence in training:} in similar form
    \item \textbf{Motivation}: The motivation for this question is similar to that of cofactor recall, focusing on the functional characterization of proteins. Enzymatic activity is a crucial aspect of protein function, requiring specific structural or sequence features for identification.
\end{itemize}

\paragraph{Location F1}
\begin{itemize}
\setlength{\itemsep}{1pt}
\setlength{\parskip}{1pt}
    \item \textbf{Question}: Where is this protein located?
    \item \textbf{Score}: F1 score
    \item \textbf{Presence in training}: in exact form
    \item \textbf{Motivation}: Proteins must be localized to specific compartments within cells to perform their functions correctly. This localization is directed by specific targeting sequences or 3D structures. Extracting this information can be challenging, and the identification of the destination for a protein can vary depending on the organism and context.
\end{itemize}

\paragraph{Binary localization average F1}
\begin{itemize}
\setlength{\itemsep}{1pt}
\setlength{\parskip}{1pt}
    \item \textbf{Question}:
    Is this a membrane protein?
    Is this a nuclear protein?
    Is this a mitochondrial protein?
    \item \textbf{Score}: Mean of F1 scores of all 3 questions
    \item \textbf{Presence in training:} never in this form
    \item \textbf{Motivation}: This question is related to the previous one but focuses on specific locations in a binary format. While the information extraction process is similar, the LLM needs to perform reasoning to find the answer.
\end{itemize}

\subsection{Metric equations used in Biochem-Lite}\label{sec:formulas}
    \begin{itemize}[leftmargin=*]
        \item \textit{mw MALE}: The mean-absolute log error (MALE) of the predicted molecular weight (MW) is:
        \vspace{2pt}
        \begin{equation}
        \setlength{\abovedisplayskip}{6pt}
        \setlength{\belowdisplayskip}{6pt}
         MALE = \frac{1}{N} \sum_{i=1}^{N} | \log_{10}(\hat{MW}_i/MW_i)|
        \end{equation}
        where $\hat{MW}_i$ is the predicted MW, $MW_i$ is the ground truth MW, and $N$ is number of examples.
        \setlength{\itemsep}{8pt}
        \item \textit{exact cofactor}: The score is computed as:
        \begin{equation}
        \setlength{\abovedisplayskip}{6pt}
        \setlength{\belowdisplayskip}{6pt}
        \text{Score}_{\text{exact}} = \frac{1}{N} \sum_{i=1}^{N} \mathbb{1}(\exists w \in R_i : w \in GT_i)
        \end{equation}
        where $R_i$ is the set of words in response $i$, $GT_i$ is the set of ground truth cofactor words for protein $i$, and $\mathbb{1}$ is the indicator function that is 1 for exactly one match and 0 otherwise.
        \item \textit{location F1}: The F1 score is the harmonic mean of Precision and Recall. The predicted class assignments for the calculation of Precision and Recall are based on the exclusive presence of correct labels (\textit{membrane}, \textit{nucleus}, or \textit{mitochondrion}) in the generated response. Responses lacking or containing multiple labels receive the \textit{none} class label.
        \item \textit{is\_enzyme F1}: The F1 score is computed similar to \textit{location F1}. 
        Responses are classified as True or False based on the exclusive presence of \textit{yes} or \textit{no}, respectively. Any deviation is labeled as \textit{none}.
        \item \textit{binary Q mean F1}: We have:
        \begin{equation}
        \setlength{\abovedisplayskip}{6pt}
        \setlength{\belowdisplayskip}{6pt}
        F1_{\text{binary Q mean}} = \frac{1}{3} \sum_{q=1}^{3} F1_q
        \end{equation}
        where $F1_q$ is the F1 score for each binary question (class assignments similar to \textit{is\_enzyme F1}).
    \end{itemize}

\subsection{MLP Baseline}\label{sec:mlp}
We use a 3-layer MLP with GELU activation on protein sequence embeddings from the Pre-trained ESM2. Where possible, we convert each light-weight benchmarking question into a classification or value prediction task. More specifically we performed regression for log values of mw for mw MALE and performed classification for \textit{is\_enzyme} and \textit{location} questions. The MLP is then trained to predict the value or correct class labels using mean squared error or cross entropy loss, respectively. This baseline sets an upper limit for the information content inherent in ESM2 embeddings.

\subsection{Training}\label{sec:training}
We devised a simple training strategy for PQA models, keeping both the PLM and LLM frozen and optimizing for the causal language model loss with AdamW. Hyperparameters were optimised following a greedy search as detailed in section \ref{sec:hyperopt}. All training were performed on a single A100-80GB or H100-80GB Nvidia GPU. Unless otherwise stated, both the QA and summary statement section of the Pika-DS was used for the training of all Pika models. Examples were split to train, validation and test set based on the UniRef50 cluster or EvoGroup of their respective protein sequences in 94.5\%, 0.05\% and 5\% ratios, respectively.

\subsection{Hyperparameter Optimization}\label{sec:hyperopt}
Greedy hyper parameter search was performed for both architectures, monitoring the is\_enzyme metric. For all experiments, training was performed on 25000 protein sequences and metrics were computed on 250 unseen proteins with gpt2-medium model as the LLM and esm2\_t12\_35M model as the protein sequence encoder. The sweep was performed, in order, on optimizer’s weight decay [0, 1e-4, 1e-2] and learning rate [1e-5, 1e-4, 1e-3], batch size [2, 4, 8], Perceiver latent size [32, 64, 100] and the number of Perceiver layers [1, 2, 4]. The final set of hyper parameters were as follows: learning rate: 1e-4, weight decay: 1e-4, batch size: 8, Perceiver latent size: 100, and number of Perceiver layers: 1 (Cross-Pika architecture), 4 (Self-Pika architecture). 

\subsection{Creation of EvoGroups}\label{sec:evogroups}

EvoGroups were decided by starting from a random sequence in the data, identifying all its related sequences using JackHammer \citep{hmmer} with a very lenient cut-off threshold for e-E-score of 1.0 to ensure a broad grouping of related proteins. For each randmoly selected sequence, all related sequences were marked as belonging to the same EvoGroup and were removed from the remainder of the dataset for further iterations with newly selected sequences. The process was repeated until all sequences were assigned an EvoGroup.


\newpage
\clearpage
\section{Supplementary Figures \& Tables}

\renewcommand{\thefigure}{\thesection.\arabic{figure}}
\setcounter{figure}{0} 
\renewcommand{\thetable}{\thesection.\arabic{table}}
\setcounter{table}{0} 

\vspace{-2pt}
\begin{figure*}[ht]
\centering
    \includegraphics[width=0.55\textwidth]{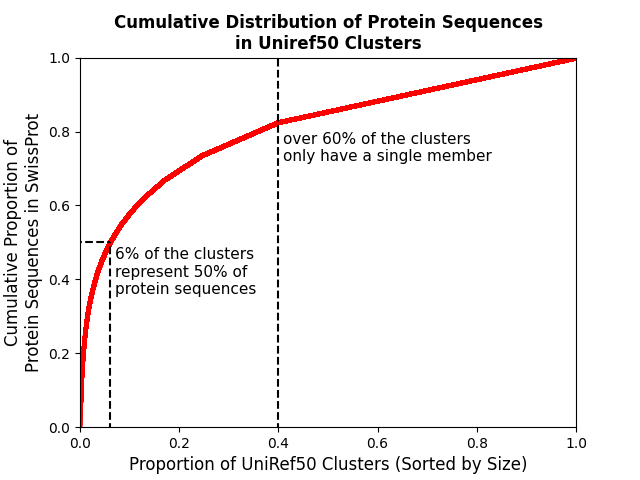}
     \captionsetup{font=small}
    \caption{Over-representation bias in SwissProt database. }\label{uniref}
\end{figure*}

\vspace{-10pt}
\begin{figure*}[ht]
\centering
    \includegraphics[width=0.7\textwidth]{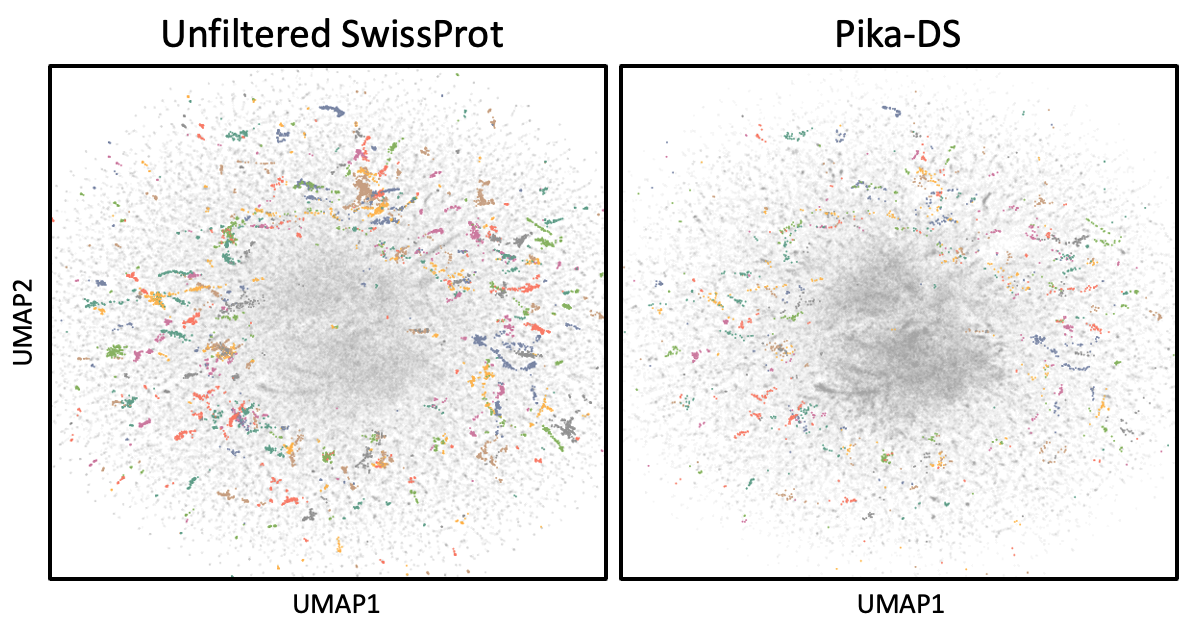}
     \captionsetup{font=small}
    \caption{Comparison of sequence bias in SwissProt database, before and after filtering. Members of the top100 largest UniRef50 clusters are colored. The strong overrepresentation of highly studied protein groups is apparent in the unfiltered plot, while the Pika-DS shows a significant reduction in this bias}\label{umaps}
\end{figure*}

\begin{figure*}[ht]
\centering
    \includegraphics[width=0.6\textwidth]{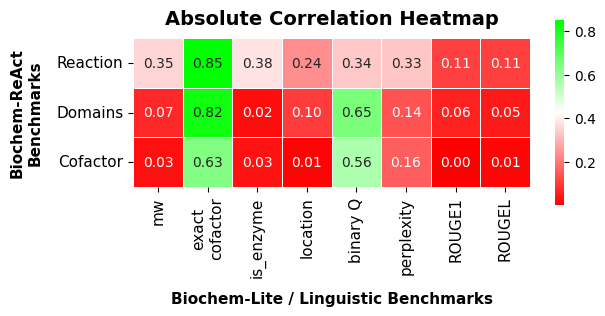}
     \captionsetup{font=small}
    \caption{Cross-correlation of Biochem-ReAct Benchmarks with Biochem-Lite/Linguistic Benchmarks. Absolute values. }\label{corrb}
\end{figure*}

\captionsetup[table]{belowskip=0pt}
{\renewcommand{\arraystretch}{1}
\begin{table}[ht]
    \centering
    \small
    \begin{tabular}{lcccc}
        \hline
        & \# total & \# unique & \# words & \# tokens \\
        \hline
        Proteins & 257 K & 252 K & - & 105.3 M \\
        Info fields & 1.17 M & 319 K & 10.8 M & 29.5 M \\
        Summaries & 1.33 M & 861 K & 22.5 M & 36.4 M \\
        Questions & 1.71 M & 98 K & 15.7 M & 14.9 M \\
        Answers & 1.71 M & 783 K & 18.4 M & 31.8 M \\
         \hline
    \end{tabular}
    \vspace{0.2cm}
    \captionsetup{justification=raggedright,singlelinecheck=false,font=small}
    \caption{Pika-DS size and token counts. Values for Summaries represent the number of sentences.}
    \setlength{\belowcaptionskip}{-10pt}
    \label{tab:dataset_count}
\end{table}}

\captionsetup[table]{belowskip=0pt}
{\renewcommand{\arraystretch}{1.3}
\begin{table*}[ht]
    \centering
    \resizebox{0.99\textwidth}{!}{
\begin{tabular}{|l|ll|}
\hline
uniref & \multicolumn{2}{l|}{UniRef50\_Q9HUN4} \\ \hline
sequence & \multicolumn{2}{l|}{MRPLVATVDLTALRHNYLLAKQCAPQRKAFAVVKANAYGHGAPEAVTALREIADGFAVACLEEAE{[}...{]}} \\ \hline
\multirow{5}{*}{fields} & \multicolumn{2}{l|}{catalytic activity: EC =   5.1.1.1, L-alanine = D-alanine} \\ \cline{2-3} 
 & \multicolumn{2}{l|}{cofactor: pyridoxal 5'-phosphate} \\ \cline{2-3} 
 & \multicolumn{2}{l|}{\begin{tabular}[c]{@{}l@{}}functional domains: alanine racemase activity,  Alanine racemase, Alanine racemase C-terminal domain-like, \\ PLP-binding barrel, pyridoxal phosphate binding\end{tabular}} \\ \cline{2-3} 
 & \multicolumn{2}{l|}{taxonomy: Bacteria, Pseudomonadota, Gammaproteobacteria} \\ \cline{2-3} 
 & \multicolumn{2}{l|}{protein size: 358 aa, 38485 KDa} \\ \hline
\multirow{5}{*}{summary} & \multicolumn{2}{l|}{\begin{tabular}[c]{@{}l@{}}This protein is an alanine racemase enzyme found in various bacteria, specifically in the \\ Pseudomonadota phylum of the Gammaproteobacteria class\end{tabular}} \\ \cline{2-3} 
 & \multicolumn{2}{l|}{It is composed of 358 amino acids with a molecular weight of 38,485 KDa} \\ \cline{2-3} 
 & \multicolumn{2}{l|}{\begin{tabular}[c]{@{}l@{}}The protein exhibits a catalytic activity with an Enzyme Commission (EC) number of 5.1.1.1, \\ where it converts L-alanine to D-alanine\end{tabular}} \\ \cline{2-3} 
 & \multicolumn{2}{l|}{Its enzymatic function is facilitated by a cofactor known as pyridoxal 5'-phosphate} \\ \cline{2-3} 
 & \multicolumn{2}{l|}{\begin{tabular}[c]{@{}l@{}}The protein contains several functional domains, including the Alanine racemase activity domain, \\ the Alanine racemase C-terminal domain-like, the PLP-binding barrel domain, and \\ the pyridoxal phosphate binding domain.\end{tabular}} \\ \hline
\multirow{8}{*}{qa} & \multicolumn{2}{l|}{\begin{tabular}[c]{@{}l@{}}What is the catalytic activity of this protein? \\ The catalytic activity of this protein is the conversion of L-alanine to D-alanine.\end{tabular}} \\ \cline{2-3} 
 & \multicolumn{2}{l|}{\begin{tabular}[c]{@{}l@{}}What is the cofactor associated with this protein? \\ The cofactor associated with this protein is pyridoxal 5'-phosphate.\end{tabular}} \\ \cline{2-3} 
 & \multicolumn{2}{l|}{\begin{tabular}[c]{@{}l@{}}Which phylum does this protein belong to? \\ This protein belongs to the Pseudomonadota phylum.\end{tabular}} \\ \cline{2-3} 
 & \multicolumn{2}{l|}{\begin{tabular}[c]{@{}l@{}}What is the molecular weight of this protein? \\ The molecular weight of this protein is 38,485 KDa.\end{tabular}} \\ \cline{2-3} 
 & \multicolumn{2}{l|}{\begin{tabular}[c]{@{}l@{}}How many amino acids are present in this protein? \\ This protein is composed of 358 amino acids.\end{tabular}} \\ \cline{2-3} 
 & \multicolumn{2}{l|}{\begin{tabular}[c]{@{}l@{}}What are the functional domains of this protein? \\ The functional domains of this protein include the Alanine racemase activity domain, \\ the Alanine racemase C-terminal domain-like, the PLP-binding barrel domain, and \\ the pyridoxal phosphate  binding domain.',\end{tabular}} \\ \cline{2-3} 
 & \multicolumn{2}{l|}{\begin{tabular}[c]{@{}l@{}}What class of bacteria is this protein found in? \\ This protein is found in the Gammaproteobacteria class.\end{tabular}} \\ \cline{2-3} 
 & \multicolumn{2}{l|}{\begin{tabular}[c]{@{}l@{}}Can this protein act as an enzyme?   \\ Yes\end{tabular}} \\ \hline
\multirow{5}{*}{metrics} & \multicolumn{1}{l|}{in\_membrane} & False \\ \cline{2-3} 
 & \multicolumn{1}{l|}{in\_nucleus} & False \\ \cline{2-3} 
 & \multicolumn{1}{l|}{in\_mitochondria} & False \\ \cline{2-3} 
 & \multicolumn{1}{l|}{is\_enzyme} & True \\ \cline{2-3} 
 & \multicolumn{1}{l|}{cofactor} & pyridoxal 5'-phosphate \\ 
 \hline
\end{tabular}}
\caption{An example entry in Pika-DS representing Uniprto ID A4VQM5}

    \label{tab:exmp}
\end{table*}}

{\renewcommand{\arraystretch}{1.2}
\begin{table*}[ht]
    \centering
    \small
    \begin{tabular}{lcccc}
        \hline
          & Random & \begin{tabular}[c]{@{}c@{}}LLM\\ only\end{tabular} & \begin{tabular}[c]{@{}c@{}}self-Pika\\ w/o PLM\end{tabular} & self-Pika \\
          \hline
        F1   Score & 0.49 & 0.00 & 0.43 & 0.99 \\
        Accuracy & 0.50 & 0.00 & 0.51 & 0.99 \\
         \hline
    \end{tabular}
    \captionsetup{font=small}
    \caption{Performance on Control Question.}
    \label{tab:controlQT}
\end{table*}}


\end{document}